%% file: main.tex
\documentclass[10pt,twocolumn,letterpaper]{article}

\usepackage[pagenumbers]{cvpr} 
\usepackage{graphicx}	
\usepackage{amsmath}	
\usepackage{amssymb}	
\usepackage{booktabs}
\usepackage{times}
\usepackage{microtype}
\usepackage{epsfig}
\usepackage{caption}
\usepackage{float}
\usepackage{placeins}
\usepackage{color, colortbl}
\usepackage{stfloats}
\usepackage{enumitem}
\usepackage{tabularx}
\usepackage{xstring}
\usepackage{multirow}
\usepackage{xspace}
\usepackage{url}
\usepackage{subcaption}
\usepackage{comment}
\usepackage[hang,flushmargin]{footmisc}

\input{preamble}
\definecolor{cvprblue}{rgb}{0.21,0.49,0.74}
\usepackage[pagebackref,breaklinks,colorlinks,allcolors=cvprblue]{hyperref}

\newcommand{\methodName}{EI-Part\xspace}

\title{EI-Part: Explode for Completion and Implode for Refinement}

\makeatletter
\let\orig@thanks\thanks
\gdef\@title@footnotes{}
\renewcommand{\thanks}[1]{%
  \orig@thanks{#1}%
  \g@addto@macro\@title@footnotes{\footnotetext[\the\c@footnote]{#1}}%
}
\makeatother
\author{
  Wanhu Sun$^{1,3}$\thanks{Work done during internship at MathMagic.}
  \quad
  Zhongjin Luo$^{1}$\footnotemark[2]
  \quad
  Heliang Zheng$^{3}$
  \quad
  Jiahao Chang$^{1,2}$
  \quad
  Chongjie Ye$^{1,2}$
  \\
  Huiang He$^{3}$
  \quad
  Shengchu Zhao$^{3}$
  \quad
  Rongfei Jia$^{3}$
  \quad
  \setcounter{footnote}{1}%
  Xiaoguang Han$^{1,2}$\thanks{Corresponding authors: Z. Luo and X. Han.}
  \\[6pt]
  $^{1}$SSE, CUHKSZ \quad $^{2}$FNii-Shenzhen \quad $^{3}$Math Magic
}
\begin{document}

\twocolumn[{
\renewcommand\twocolumn[1][]{#1}%
\maketitle
\begin{center}
  \newcommand{\teaserwidth}{\textwidth}
  \centerline{
    \includegraphics[width=1.\textwidth]{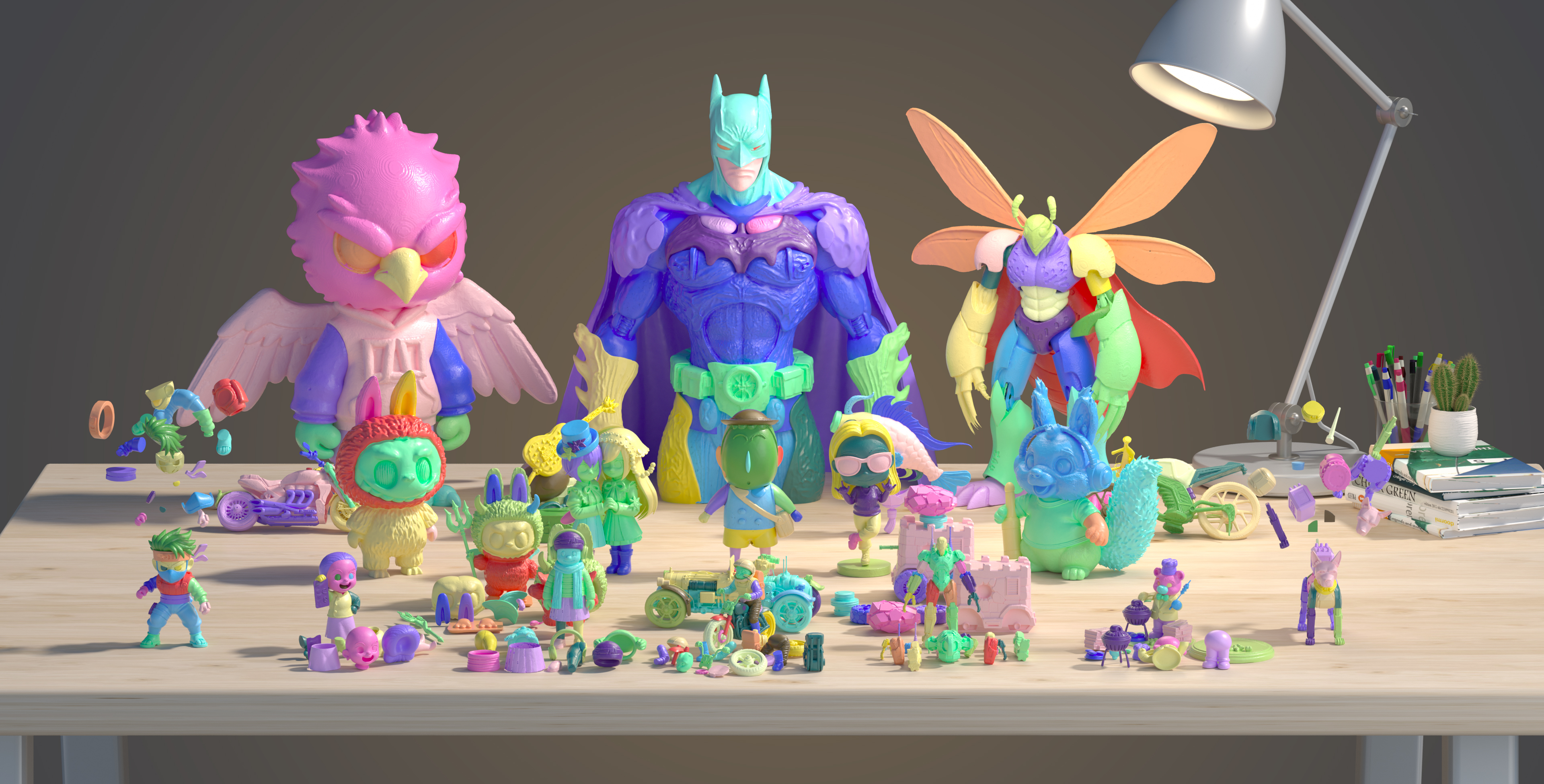}}
    \captionof{figure}{We present \methodName, a novel framework designed to efficiently produce structurally coherent and semantically meaningful parts with fine-grained geometric details. Our work proposes an Explode-Implode strategy to fully utilize spatial resolution at different stages for meaningful part completion and geometric detail generation.}
  \label{fig:teaser}
\end{center}
}]

\makeatletter
\footnotetext[1]{Work done during internship at MathMagic.}
\footnotetext[2]{Corresponding authors: Z. Luo and X. Han.}
\makeatother   
\input{section/00_abstract}    
\input{section/01_intro}
\input{section/02_related}

\input{section/03_method}
\input{section/04_experiments}
\input{section/05_conclusion}

\clearpage

{
    \small
    \bibliographystyle{ieeenat_fullname}
    \bibliography{main}
}

\input{section/X_suppl}

\end{document}

%% file: section/00_abstract.tex
\begin{abstract}
Part-level 3D generation is crucial for various downstream applications, including gaming, film production, and industrial design. However, decomposing a 3D shape into geometrically plausible and meaningful components remains a significant challenge. Previous part-based generation methods often struggle to produce well-constructed parts, exhibiting either poor structural coherence, geometric implausibility, inaccuracy, or inefficiency. 
To address these challenges, we introduce \methodName, a novel framework specifically designed to generate high-quality 3D shapes with components, characterized by strong structural coherence, geometric plausibility, geometric fidelity, and generation efficiency. We propose utilizing distinct representations at different stages: an Explode state for part completion and an Implode state for geometry refinement. This strategy allows us to fully leverage spatial resolution, enabling flexible part completion and fine geometric details generation. To maintain structural coherence between parts, a self-attention mechanism is incorporated in both the exploded and imploded states, facilitating effective information perception and feature fusion among components during generation.
Extensive experiments conducted on various benchmarks demonstrate that \methodName efficiently yields semantically meaningful and structurally coherent parts with fine-grained geometric details, achieving state-of-the-art performance in part-level generation compared to existing methods. \noindent\textbf{Project page:} \url{https://cvhadessun.github.io/EI-Part/}
\end{abstract}

%% file: section/01_intro.tex
\section{Introduction}
\label{sec:introduction}
3D content generation has become increasingly vital across various fields, including gaming, film production, and industrial design. Recent advancements in this area have been driven by the emergence of extensive 3D datasets~\cite{objaverse,objaverseXL} and sophisticated generative models~\cite{rombach2022high,liu2023zero,zhang2024clay,xiang2025structured}. However, despite these advancements, many current methods~\cite{li2024craftsman3d,zhao2025hunyuan3d,li2025triposg,ye2025hi3dgen,chen2025ultra3d,li2025sparc3d} produce 3D assets in a monolithic form, lacking the nuanced part-level decomposition essential for detailed manipulation and editing. This limitation significantly hinders their ability to create complex assets that can be easily customized for various applications.

Within this landscape, part-based generation plays a critical role in the overall process of 3D content creation. To ensure that the generated parts meet downstream functionality, achieving high-quality part-level generation poses several essential criteria: 1) \textbf{Structural Coherence.} Maintaining structural coherence among parts is vital to ensure that the relationships and connections between different components are logical and consistent. 2) \textbf{Geometric Plausibility.} The generated parts must be geometrically plausible, meaning that each component should fit well within the overall structure and possess meaningful shapes and features that align with its intended function. 3) \textbf{Fidelity.} Achieving high fidelity in geometric details is essential for producing realistic outputs that accurately capture both visual realism and fine geometric characteristics of real-world objects. 4) \textbf{Efficiency.} Enhancing computational efficiency in both time and space is crucial for reducing the costs associated with part generation, thereby facilitating faster iterations and adjustments during the 3D design process. Coherence and plausibility are crucial for functional performance, while fidelity is essential for aesthetic appeal, and efficiency is important for cost reduction, all of which jointly contribute to effective 3D part generation.

Although several works~\cite{yang2025holopart,zhang2025bang,lin2025partcrafter,tang2024partpacker,yan2025x,yang2025omnipart} have emerged to focus on 3D part generation, many of these efforts do not fully consider the aforementioned criteria, leading to failures to produce well-constructed parts.
BANG~\cite{zhang2025bang} and PartPacker~\cite{tang2024partpacker} embed all parts into either a single set or a dual set of latent codes, which constrains the representation capability of each part and consequently results in poor geometric details. HoloPart~\cite{yang2025holopart} enhances geometric details by representing each part as an independent set of latent codes and generating each individual part independently, but it neglects the relationships between parts, leading to poor structural coherence among them. PartCrafter~\cite{lin2025partcrafter} and X-Part~\cite{yan2025x} enhance structural coherence by incorporating a cross-attention mechanism among part tokens during the generation process. However, their reliance on fixed-length part tokens is inflexible for parts with varying sizes, which results in inefficient resource allocation (limited capability to represent large components and wasted capacity for small components). OmniPart~\cite{yang2025omnipart} improves representation efficiency by utilizing voxel-based representations, which adaptively allocate spatial resolution to each part and generate part shapes within each designated region defined by bounding boxes. However, assigning voxels in their consolidated state often leads to two issues: 1) overlapping voxels between parts, which would result in ambiguity during the completion process, and 2) can only occur within the limited active voxels confined by the bounding boxes, preventing any expansion beyond these boundaries. These two issues may jointly hinder its ability to produce meaningful parts. In conclusion, all existing methods struggle to produce well-constructed parts, exhibiting either poor structural coherence (e.g., HoloPart), geometric implausibility (e.g., OmniPart), inaccuracy (e.g., BANG, PartPacker), or inefficiency (e.g., PartCrafter, X-Part).

In this paper, we introduce \methodName, a novel framework designed to generate well-constructed parts through an \textit{Explode for Completion and Implode for Refinement} strategy. Our work meets the aforementioned criteria through two primary strategies: \textit{1) Exploding incomplete 3D parts to fully utilize spatial capacity for meaningful part generation, and 2) Imploding coarse part shapes to maximize spatial resolution for effectively capturing fine-grained geometric details}. Specifically, we first repurpose a multi-view diffusion model for mesh segmentation to obtain the initial incomplete 3D parts. Next, to enhance efficiency and geometric plausibility, we represent these incomplete parts as sparse voxels and apply an explosion strategy to complete the parts in a dispersed state. This explosion strategy not only avoids ambiguity caused by overlapping voxels between parts but also expands the resolution available for part completion, allowing for the generation of reasonable and meaningful components. Finally, to improve geometric fidelity, we propose reverting the exploded 3D voxels to an imploded state, fully utilizing spatial resolution to refine geometric details. Additionally, to maintain structural coherence between parts, we adopt a self-attention mechanism in both the exploded and imploded states, enabling information perception and feature fusion among components during shape completion and geometry refinement. 

To evaluate the effectiveness of \methodName, we conducted extensive experiments on various benchmarks. Our results demonstrate that \methodName can efficiently generates semantically meaningful and structurally coherent parts with fine-grained geometries. Comparisons with existing approaches show that \methodName achieves state-of-the-art performance in part-level generation. In summary, the contributions of our work are as follows:
\begin{itemize}
\item We propose \methodName, a novel framework dedicated to producing high-quality 3D shapes with well-constructed parts, characterized by strong structural coherence, geometric plausibility, accuracy, and efficiency.

\item We introduce an Explode-Implode strategy to fully utilize spatial resolution at different stages: Explode for meaningful part completion and Implode for geometric details refinement.

\item Extensive experiments demonstrate that our method achieves superior performance in part-level generation compared to state-of-the-art approaches.
\end{itemize}

%% file: section/02_related.tex
\section{Related Work}
\label{sec:related_work}

\paragraph{3D Object Generation.} 
Traditionally, the limitations of data scale and computational resources have constrained 3D generation and reconstruction to specific categories~\cite{chang2015shapenet,wang2018pixel2mesh,park2019deepsdf,mescheder2019occupancy,luo2023sketchmetaface,saito2019pifu}. In the past few years, the emergence of large 3D datasets~\cite{objaverse,objaverseXL}, efficient representations~\cite{zhang20233dshape2vecset,shue20233d,ren2024xcube,kerbl20233d,li2023diffusion}, and advanced generative models~\cite{rombach2022high,liu2023zero,zhang2024clay,xiang2025structured} has facilitated a significant evolution in 3D object generation, transitioning from category-specific approaches to open-world object generation. 

As a pioneering effort, CLAY~\cite{zhang2024clay} integrates large-scale data with sophisticated transformer-based architectures for direct 3D generation. Subsequently, many works have aimed to enhance accuracy, efficiency, and controllability. TripoSG~\cite{li2025triposg} leverages large-scale rectified flow transformers for high-fidelity 3D shape synthesis. HunYuan3D~\cite{zhao2025hunyuan3d} employs scalable flow-based diffusion transformers alongside with high-quality datasets to create geometries that align with input conditions. TRELLIS~\cite{xiang2025structured} utilizes a voxel-based structured latent representation to achieve precise geometry generation in a coarse-to-fine manner. Hi3DGen~\cite{ye2025hi3dgen} proposes a method to generate high-fidelity 3D geometry from images by utilizing normal maps as a bridge. Sparc3D~\cite{li2025sparc3d} introduces a sparse deformable marching cubes representation for high-resolution generative modeling, facilitating the generation of fine-grained shape details.  Furthermore, some recent works explore the generation of artist-like topologies and leverage autoregressive models to directly produce explicit faces~\cite{siddiqui2024meshgpt,chen2024meshanything,hao2024meshtron,wang2025nautilus}. However, the aforementioned methods typically generate 3D objects as monolithic entities without part decomposition, which limits subsequent manipulation, editing and animation. Our work proposes a novel method for efficiently generating well-constructed parts, thereby facilitating downstream applications.

\paragraph{3D Part Segmentation.}
Traditional methods primarily extract high-level point features and predict per-point labels for 3D segmentation~\cite{zhao2021point,qi2017pointnet++,qi2017pointnet}. Most of these approaches rely on extensive annotations for supervised learning, which limits them to category-specific segmentation and hinders their ability to generalize to open-world object scenarios. With the advent of 2D foundation models~\cite{kirillov2023segany,hafner2021clip,caron2021emerging,oquab2023dinov2}, some methods have begun to lift 2D visual knowledge to 3D domains~\cite{umam2023partdistill,zhong2024meshsegmenter,xu2025sampro3d,tang2024segment}. While these approaches can generalize to open-world objects, their segmentation performance remains constrained, particularly in addressing self-occlusion scenarios. 

In recent years, several works have aimed to leverage large-scale 3D datasets and advanced diffusion models to enhance the generalization of 3D segmentation. SAMPart3D~\cite{yang2024sampart3d} introduces a scalable zero-shot 3D part segmentation approach that distills part-level 2D visual features into the 3D domain using a trained 3D backbone for multi-granularity segmentation. PartField~\cite{liu2025partfield} proposes a feedforward framework that learns a part-based 3D feature field through contrastive learning and performs part decomposition via clustering. P3-SAM~\cite{ma2025p3} presents a 3D-native point-promptable part segmentation model that integrates point features across different levels, leveraging point-wise features for 3D part segmentation. However, due to limitations in 3D resolution, these methods often yield unsatisfactory segmentation results and struggle to produce clear boundaries. In this work, we propose a multi-view diffusion model for part segmentation to achieve robust segmentation results with well-defined boundaries.

\paragraph{3D Part Generation.}
Beyond 3D part segmentation, 3D part generation seeks to create complete, geometrically plausible, and meaningful components that fulfill practical requirements in both aesthetic appeal and intended functionality. However, existing methods struggle to produce well-constructed parts. PartGen~\cite{chen2024partgen} reconstructs components from multi-view segmented results independently, resulting in poor structural coherence. 

BANG~\cite{zhang2025bang} formulates part generation as an object explosion process but embeds parts into a single set of latent codes, leading to insufficient geometric detail. Similarly, the dual latent codes used in PartPacker~\cite{tang2024partpacker} restrict its ability to capture fine geometric details. HoloPart~\cite{yang2025holopart} generates each part independently, neglecting the interrelationships between parts and thereby diminishing structural coherence. To enhance coherence, PartCrafter~\cite{lin2025partcrafter} incorporates cross-attention mechanisms between parts, while X-Part~\cite{yan2025x} introduces an inter-part attention module. However, their reliance on fixed-length part tokens is inflexible for addressing parts of varying sizes, resulting in inefficient resource allocation. OmniPart~\cite{yang2025omnipart} employs voxel-based representations to adaptively allocate space for components of different sizes. Nonetheless, the use of bounding boxes not only restricts the available resolution for completion but also leads to overlapping voxels between parts, thereby hindering the generation of meaningful components. 

In this paper, we propose a novel framework designed to produce high-quality 3D shapes with well-constructed parts, achieving strong structural coherence, geometric plausibility, accuracy, and efficiency.

%% file: section/03_method.tex
\section{Method}
\label{sec:method}

\input{figure/fig_pipeline}

The overview of our method is shown in Fig.~\ref{fig:pipeline}. Given an input 3D shape, we first obtain its normal map and canonical coordinate maps from six views. We then perform frontal segmentation using SAM and employ MVSegNet to achieve multi-view consistent segmentations. These segmentations are lifted to 3D to create an initial part segmentation, which is subsequently inpainted by InSegNet to produce accurate 3D segmented shapes. The segmented result is exploded into discrete 3D voxels, enabling us to perform conditional diffusion completion to generate geometrically plausible and complete part structures. Next, the exploded complete parts are imploded back to a compact state for conditional diffusion refinement, capturing fine-level details. Thanks to the utilization of the exploded-imploded strategy, we fully utilize spatial resolution at different stages, enhancing the accuracy of part completion and surface details. As a result, the 3D shapes generated by our method exhibit individual, structurally consistent, and geometrically plausible part shapes with fine-grained geometric details.

\subsection{Diffusion Part Segmentation}
Given an input 3D shape, we first decompose it into semantically meaningful parts. Inspired by texturing techniques~\cite{bensadoun2024meta, zhao2025hunyuan3d, yang2025pandora3d, liang2025unitex}, we adopt a two-stage diffusion-based framework for robust 3D part segmentation. This framework first generates multiview segmentations and then propagates this semantic knowledge into 3D space to learn a continuous semantic field.

To be specific, for an input shape $O$, we render its normal maps $\{n_i\}_{i=1}^6$ and canonical coordinate maps (CCMs) $\{c_i\}_{i=1}^6$ from six orthographic views. We then obtain a frontal-view segmentation $s_1$ using SAM~\cite{kirillov2023segany}. The normal maps $\{n_i\}_{i=1}^6$, CCMs $\{c_i\}_{i=1}^6$, and frontal segmentation $s_1$ serve as input for our multiview segmentation diffusion model (MVSegNet) to generate consistent segmentations across six viewpoints. A straightforward approach for obtaining 3D segmentation is to directly unproject the multiview segmentations onto the 3D shape. However, this method would lead to inaccuracies in occluded and unseen regions. To address this, we utilize a  inpainting model (InSegNet) to refine part segmentation. Taking the multiview segmentations and the unprojected 3D segmentations as input, InSegNet extracts semantic features $(f_{2D}, f_{3D})$ from both 2D and 3D segmentation, subsequently learning a continuous 3D semantic field. Formally, for each point $x$ on the 3D surface, we can query its semantic color from the learned field via $\hat{S}(x) = \operatorname{InSegNet}(x, f_{2D}, f_{3D})$. The InSegNet is trained by minimizing the following loss function~\cite{saito2019pifu,liang2025unitex},
\begin{equation}
\mathcal{L}_{seg} = \frac{1}{n} \sum_{i=1}^n |\hat{S}(x_i) - S(x_i)|,
\end{equation}
where $\hat{S}(\cdot)$ denotes the predicted semantic RGB color and $S(\cdot)$ represents the ground truth. $x_i$ denotes the $i$-th sampled point, and $n$ is the number of sampled points. Given the learned semantic field, we densely sample points from $O$ and query their semantic colors, obtaining $K$ globally consistent and seamless 3D parts $\{p_{s}^k\}_{k=1}^K$.

\subsection{Explode for Completion}
Although the 3D segmentation outputs individual parts, their incompleteness limits their applicability in downstream tasks. To achieve geometrically plausible and complete part structures for their intended functionality, we propose to explode the 3D segmented points into a dispersed state, thereby extending the available spatial resolution for better part completion. Compared to OmniPart~\cite{yang2025omnipart}, which is limited to performing completion within active voxels, our part completion strategy fully leverages spatial resolution to expand available completion extents.

Specifically, we convert $\{p_{s}^k\}_{k=1}^K$ into explicit segmented voxels $\{v_{s}^k\}_{k=1}^K$ and utilize a sparse structured representation for subsequent part generation. The voxel-based representation enables us to flexibly represent parts of varying sizes, with larger parts occupying more voxels and smaller parts fewer, thereby enhancing spatial efficiency. Inspired by~\cite{zhang2025bang}, we employ explosion vector optimization to explode $\{v_{s}^k\}_{k=1}^K$  radially outward. We compute the axis-aligned bounding boxes for each part $v_{s}^k$ and optimize a translation vector to diffuse $\{v_{s}^k\}_{k=1}^K$ from a converged state into a dispersed state $e_s = \operatorname{Explode}(\{v_{s}^k\}_{k=1}^K)$. The translation directions $\{u_k\}_{k=1}^K$ and distances $\{\mathbf{d}_k\}_{k=1}^K$ are recorded to facilitate subsequent implosion.
Let $e_c$ be complete parts, given the frontal normal map $n_1$ and the exploded part voxels $e_s$ as input, we model the conditional part completion as a structured diffusion process $p_\theta(e_c \mid e_s, n_1)$. Thanks to the utilization of exploded voxels, structural information perception between part voxels, as well as cross-modal feature fusion between the normal map and the voxels, can be integrated in a highly efficient manner~\cite{batifol2025flux} during the completion process:
\begin{equation}
\mathcal{F}_{E} = \operatorname{CrossAttn}(D_n, \operatorname{SelfAttn}(\operatorname{Concat}(E_s, Z_t))).
\end{equation}
These operations are performed in latent feature space. We obtain 2D tokens $D_n$ of the normal map using DINOv2~\cite{oquab2023dinov2} and 3D voxel-based conditional latent features $E_s$ of the incomplete parts via the SS-VAE encoder~\cite{xiang2025structured} to generate the target structured latents $E_c$ of the complete parts. $Z_t$ denotes the latent noise at the $t$-th step. This advantageous fusion of different features significantly enhances both the structural coherence and accuracy of the resulting complete parts. Our conditional completion is formulated as a rectified flow-based diffusion process with DiT~\cite{peebles2023scalable} and trained using Conditional Flow Matching (CFM)~\cite{lipman2022flow}:
\begin{equation}
\mathcal{L}_{CFM}(\theta)=\mathbb{E}_{\boldsymbol{x}_0, t, \boldsymbol{\epsilon}}\left\|\boldsymbol{v}_\theta(\boldsymbol{x}, t)-\left(\boldsymbol{\epsilon}-\boldsymbol{x}_0\right)\right\|_2^2 .
\label{eq:cfm_loss}
\end{equation}

\subsection{Implode for Refinement}
In the completion stage, we represent incomplete part voxels in an exploded state to generate complete part structures. This process is designed to fully utilize spatial resolution to produce complete, structurally coherent, and geometrically plausible parts while disregarding fine-level geometric details. Consequently, we propose to implode the complete exploded coarse part structures into an imploded state, thereby maximizing spatial resolution again to capture fine-grained geometric details.

In particular, the generated structured latents $E_c$ is decoded to obtain the complete part voxels $e_c$. To implode $e_c$ into a more compact spatial distribution, we sort all part voxels by their distance from the center and adjust their positions. Instead of moving parts along the opposite direction $\{-u_k\}_{k=1}^K$ of the explosion by the exploded distance $\{\mathbf{d}_k\}_{k=1}^K$ via $m_k^{\prime} = m_k - \mathbf{d}_k \cdot u_k$, we propose to update the positions $\{m_k\}_{k=1}^K$ of the parts iteratively by,
\begin{equation}
m_{k}^{j+1} = m_{k}^{j} - \alpha \cdot u_k,
\end{equation}
where $\alpha$ is the step size that determines how quickly the parts approach the center, while $j$ denotes the $j$-th iteration. This iterative process would stop when collisions occur among the parts.
After obtaining the imploded complete voxels $g_c$, we model the geometric detail refinement as a conditional generative diffusion process. Let $g_d$ represent high-fidelity parts with fine-grained geometric details. Given the frontal normal map $n_1$, the exploded and imploded complete voxels ($e_c$, $g_c$), the refinement process is formulated as $p_\theta(g_d \mid g_c, e_c, n_1)$. Similarly, we integrate structural information and cross-modal features by:
\begin{equation}
\mathcal{F}_{I} = \operatorname{CrossAttn}(D_n, \operatorname{SelfAttn}(\operatorname{Concat}(G_c, E_c, Z_t))).
\end{equation}
The refinement process is also executed in structured latent space, and the model is trained using the CFM loss as described in Eq.~\ref{eq:cfm_loss}. The 2D tokens $D_n$ are also extracted using DINOv2. In contrast to the completion stage, we utilize a more powerful VAE, Sparc3D~\cite{li2025sparc3d}, to extract voxel-based conditional features from ($G_c$, $E_c$) and decode the target part geometries $g_d$ from the generated structured latents $G_d$. As a result, $g_d$ contains individual, structurally consistent and geometrically plausible part shapes with fine-grained geometric details. 

%% file: figure/fig_pipeline.tex
\begin{figure*}[htbp]
  \centering
  \includegraphics[width=1.\linewidth]{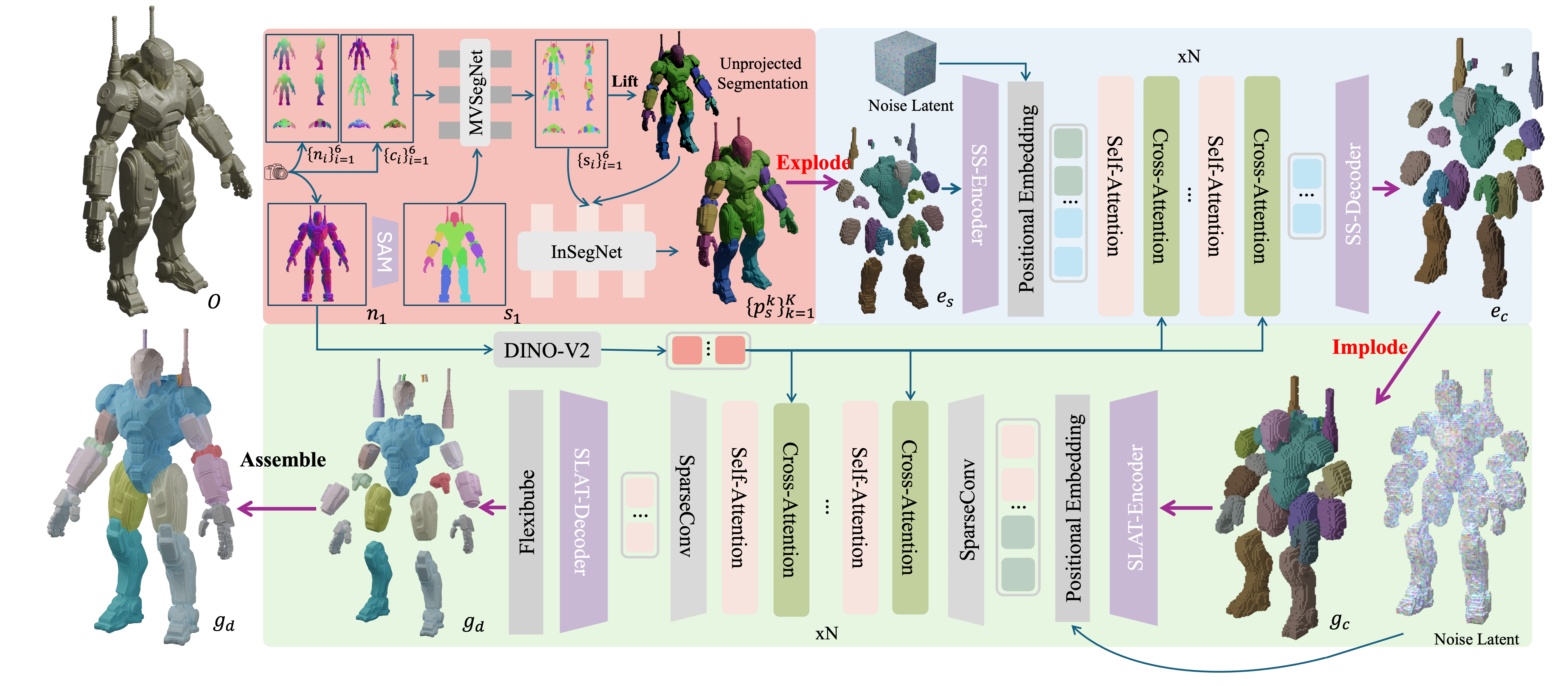} 
  \caption{The pipeline of our proposed \methodName. Given an input 3D shape $O$, we first obtain its normal map $\{n_i\}_{i=1}^6$ and canonical coordinate maps (CCMs) $\{c_i\}_{i=1}^6$ from six views. We then perform frontal segmentation using SAM and employ MVSegNet to achieve multi-view consistent segmentations. These segmentations are lifted to 3D to create an initial part segmentation, which is subsequently inpainted by InSegNet to produce accurate 3D segmented shapes $\{p_{s}^k\}_{k=1}^K$. The segmented result is exploded into discrete 3D voxels, enabling us to perform conditional diffusion completion to generate geometrically plausible and complete part structures $e_c$. Next, the exploded complete parts are imploded back to a compact state for conditional diffusion refinement, capturing fine-level details $g_d$. Thanks to the utilization of the exploded-imploded strategy, we fully utilize spatial resolution at different stages, enhancing the accuracy of part completion and surface details. As a result, the 3D shapes $g_d$ generated by our method exhibit individual, structurally consistent, and geometrically plausible part shapes with fine-grained geometric details.}
  \label{fig:pipeline}
\end{figure*}


%% file: section/04_experiments.tex
\section{Experiments}
\label{sec:experiments}

\input{table/table_compare}
\input{figure/fig_compare}
\input{figure/fig_compare_2}

\paragraph{Data Preparation.}
We collected datasets from well-known 3D object datasets including Objaverse~\cite{objaverse}, Objaverse-XL~\cite{objaverseXL}, ABO~\cite{collins2022abo}, 3D-FUTURE~\cite{fu20213d} and HSSD~\cite{khanna2024habitat}. The GLB files in these datasets are 3D models created by professionals, some of which inherently possess individual parts. We adopt the following strategy to obtain our data: 1) Filtering. We establish a threshold of 20 parts to construct the initial dataset, filtering out GLB files with excessive part counts. 2) Processing. For GLB files in the initial dataset, we further split the files based on connectivity, resulting in numerous independent sub-meshes. When the number of sub-meshes exceeds 20, we sort them by face count, surface area, and a weighted bounding box value. The n smallest sub-meshes are then merged into larger part meshes based on collision proximity (or the nearest bbox if no collision exists) to construct the final GLB.

\paragraph{Implementation details.} We trained our model utilizing the collected data. Both MVSegNet and InSegNet were optimized using the Adam optimizer, with a learning rate of $3 \times 10^{-4}$, over a duration of 4 days on 32 GPUs. For the exploded completion model, we employed the Adam optimizer with a learning rate of $1 \times 10^{-4}$, incorporating gradient clipping with a maximum norm of $1.0$ to enhance training stability. This training was conducted on 64 GPUs for a total of two weeks (14 days), with a probability of $0.3$ for introducing zero conditions to further improve robustness. The imploded refinement model was fine-tuned using the Adam optimizer at a learning rate of $1 \times 10^{-4}$ on the Sparc3D checkpoint, with a training duration of 6 days on 64 GPUs. Please refer to the supplementary materials for additional implementation details.

\paragraph{Evaluation Metrics.} 
We utilize Chamfer Distance (CD), Intersection over Union (IoU), and F-Score to assess the accuracy and completeness of the generated geometry. CD, IoU, and F-Score are evaluated by sampling 100,000 points from the 3D outputs, with all object points normalized to the range of $[-1,1]^3$. When calculating the F-Score, the radius is set to 0.1, 0.05, and 0.01, while the Voxel F-Score is set to 0.01. Voxel F-Score and Voxel IoU are computed using 3D voxels. The evaluation is conducted using PartVerse~\cite{dong2025copart}.

\paragraph{Comparison with Baseline Methods.}
To validate the effectiveness of our approach, we compare it against state-of-the-art part generation models, both quantitatively and qualitatively: BANG~\cite{zhang2025bang}, HoloPart~\cite{yang2025holopart}, X-Part~\cite{yan2025x}, and OmniPart~\cite{yang2025omnipart}. Tab.~\ref{tab:compare} shows the quantitative comparisons. Our proposed method achieves the best scores against the baselines. Fig.~\ref{fig:compare} provides qualitative results of 3D part generation. Please refer to our supplementary materials for a qualitative comparison with additional methods (e.g., PartPacker~\cite{tang2024partpacker} and PartCrafter~\cite{lin2025partcrafter}). As observed, BANG~\cite{zhang2025bang} struggles to capture fine geometric details that align with the input model. This issue arises from its practice of embedding all components into a single set of latent tokens, which restricts the representational capacity of each individual part and ultimately leads to inaccuracies in geometric details. The inaccurate 3D part segmentation employed in HoloPart~\cite{yang2025holopart}, X-Part~\cite{yan2025x}, and OmniPart~\cite{yang2025omnipart} results in numerous implausible parts. HoloPart~\cite{yang2025holopart} enhances geometric details by representing each part as an independent set of latent codes and generating them individually; however, it neglects the interrelationships between parts, resulting in unnatural completion outcomes akin to merely filling a hole, thereby leading to poor structural coherence. Meanwhile, X-Part~\cite{yan2025x} improves structural coherence through a cross-attention mechanism among parts, yet its reliance on fixed-length part tokens is inflexible for components of varying sizes. This limitation results in inefficient resource allocation, yielding insufficient capability to represent larger components while wasting capacity on smaller ones. While OmniPart~\cite{yang2025omnipart} utilizes voxel-based representations to adaptively allocate spatial resolution to different parts, it only supports completion within active voxels, thereby restricting the available spatial extents for completion. This limitation can lead to implausible part geometry and decreased geometric fidelity. Our method demonstrates proficiency in generating high-quality 3D shapes with well-defined parts, achieving notable performance in structural coherence, geometric plausibility, geometric fidelity, and efficiency. Notably, the Explode-Implode strategy enables our approach to robustly perform meaningful part completion and fine-level geometric generation. To evaluate the impact of our proposed Explode-Implode generation framework, we conduct a comparison with segmentation-based methods using the same 3D segmentation (our segmentation) as input. Segmentation-based methods refer to those approaches that take 3D segmentation as input for part generation, including HoloPart, X-Part, OmniPart and our EI-Part. As shown in Fig~\ref{fig:compare_2}, despite using the same segmentation, our method still demonstrates superior performance in both geometric plausibility and fidelity.

\input{figure/fig_ablation_EI}

\paragraph{Ablation Study on Explode and Implode.} 
To demonstrate the necessity of our Explode and Implode strategy, we conduct an ablation study in the following scenarios: 1) without both explosion and implosion; 2) with explosion but without implosion; and 3) with both components (i.e., our proposed method). As illustrated in Fig.~\ref{fig:abl_EI}, the results generated without explosion yield implausible part geometry, while the results generated without implosion fail to capture fine geometric details. The corresponding quantitative evaluations can be found in the supplementary material.

\paragraph{Ablation Study on Part Segmentation.}
To highlight the importance of our part segmentation model, we conduct an ablation study comparing our model with alternative methods (i.e., PartField~\cite{liu2025partfield} and P3-SAM~\cite{ma2025p3}). As shown in Fig.~\ref{fig:abl_PS}, the results produced by our method exhibit more accurate and meaningful segmentation.

\input{figure/fig_ablation_PS}


%% file: table/table_compare.tex
\begin{table*}[htbp]
    \centering
    \caption{Quantitative comparison between our method and state-of-the-art methods in part generation. Our method achieves the best performance across all evaluation metrics, demonstrating superior shape fidelity and geometric accuracy. Note that BANG is a commercial product with no open-source code available, making quantitative comparison infeasible.}
    \resizebox{0.68\linewidth}{!}{
        \begin{tabular}{l|cccccc}
          \textbf{Method} & \textbf{Voxel IOU} $\uparrow$ & \textbf{CD} $\downarrow$ & \textbf{Voxel F-Score 0.01} $\uparrow$ & \textbf{F-Score 0.1} $\uparrow$ & \textbf{F-Score 0.05} $\uparrow$ & \textbf{F-Score 0.01} $\uparrow$ \\
        \midrule
        PartPacker~\cite{tang2024partpacker}   & 0.2586 & 0.1273 & 0.3768 & 0.8199 & 0.6428 & 0.2435  \\
        PartCrafter~\cite{lin2025partcrafter}  & 0.0742 & 0.3474 & 0.1316 & 0.4429 & 0.2801 & 0.0749 \\
        HoloPart~\cite{yang2025holopart}     & 0.6106 & 0.0431 & 0.7374 & 0.9557 & 0.9402 & 0.6400 \\
        X-Part~\cite{yan2025x} & 0.7478 & 0.0599 & 0.8413 & 0.9256 & 0.9087 & 0.7923 \\
        OmniPart~\cite{yang2025omnipart} & 0.2861 & 0.1431 & 0.4007 & 0.7911 & 0.6516 & 0.2644 \\
        \textbf{Ours}         & \textbf{0.7981}  & \textbf{0.0194} & \textbf{0.8452} & \textbf{0.9910} & \textbf{0.9742} & \textbf{0.8129} \\
        \end{tabular}
    }
    \label{tab:compare}
\end{table*}

%% file: figure/fig_compare.tex
\begin{figure*}[htbp]
  \centering
  \includegraphics[width=.98\linewidth]{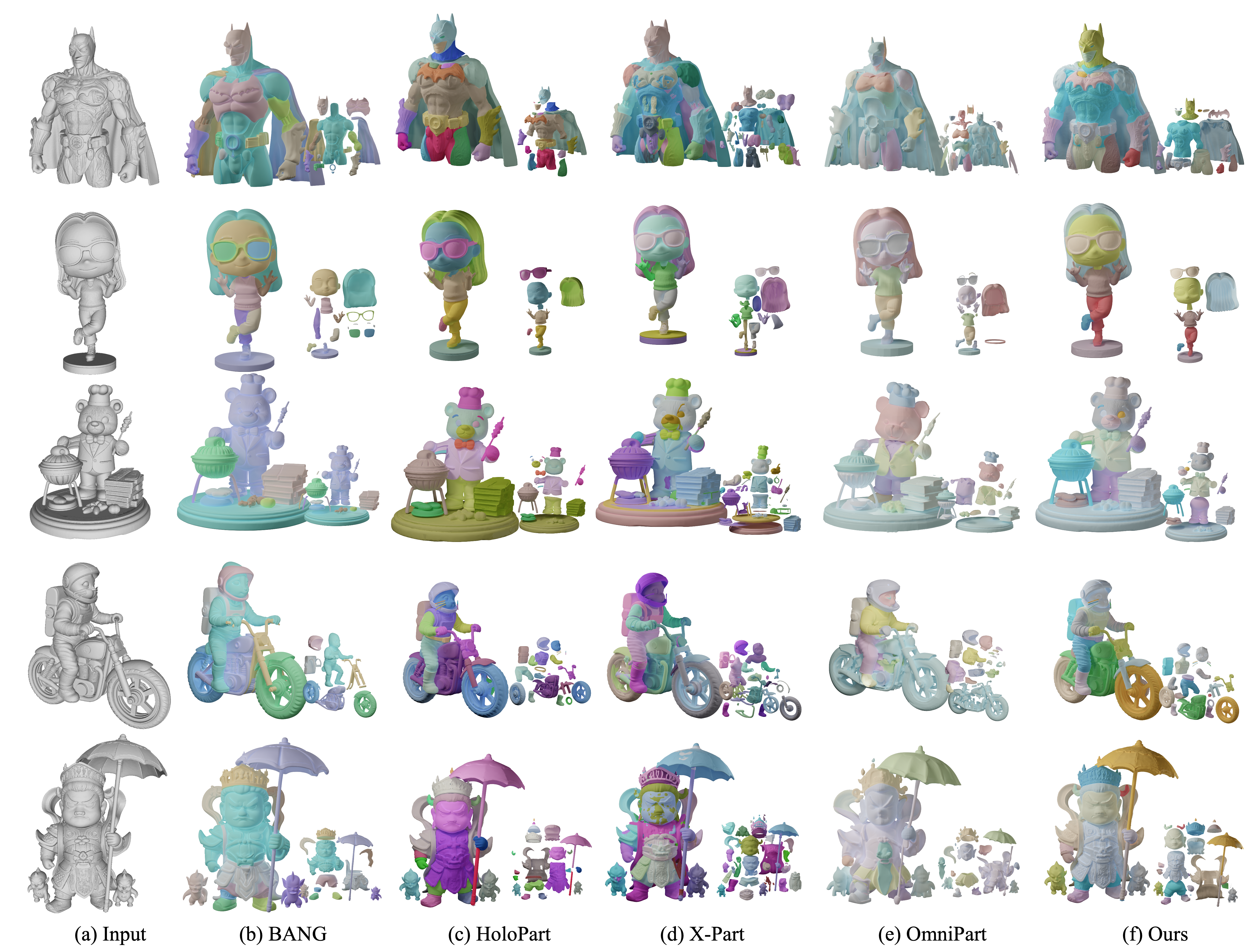}
  \caption{Qualitative comparison between ours and the state of the arts. For each row, the input model (a) is followed by the parts generated by (b) BANG~\cite{zhang2025bang}, (c) HoloPart~\cite{yang2025holopart}, (d) X-Part~\cite{yan2025x}, (e) OmniPart~\cite{yang2025omnipart} and (f) our method. Our method demonstrates proficiency in generating high-quality 3D shapes with well-defined parts, achieving notable performance in structural coherence, geometric plausibility, fidelity, and efficiency. Please refer to our supplementary materials for a qualitative comparison with additional methods (e.g., PartPacker and PartCrafter).}
  \label{fig:compare}
\end{figure*}

%% file: figure/fig_compare_2.tex
\begin{figure*}[htbp]
  \centering
  \includegraphics[width=.88\linewidth]{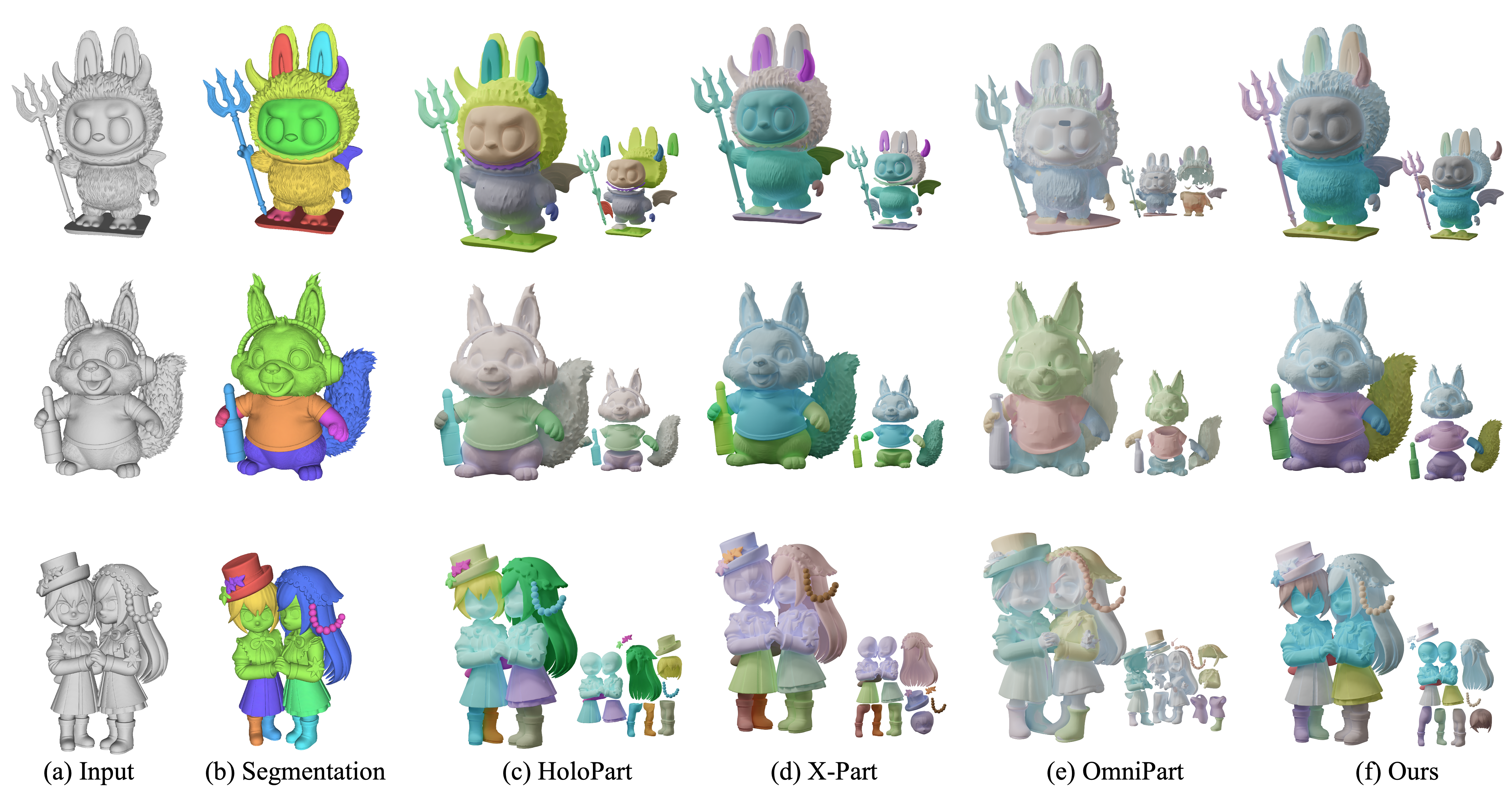}
  \caption{Qualitative comparison between our method and the state of the art, where all methods share the segmentation results of ours. Each row presents the input model (a), followed by (b) our segmentation, and the parts generated by (c) HoloPart~\cite{yang2025holopart}, (d) X-Part~\cite{yan2025x}, (e) OmniPart~\cite{yang2025omnipart} and (f) our method.}
  \label{fig:compare_2}
\end{figure*}

%% file: figure/fig_ablation_EI.tex
\begin{figure}[htbp]
  \centering
  \includegraphics[width=0.98\linewidth]{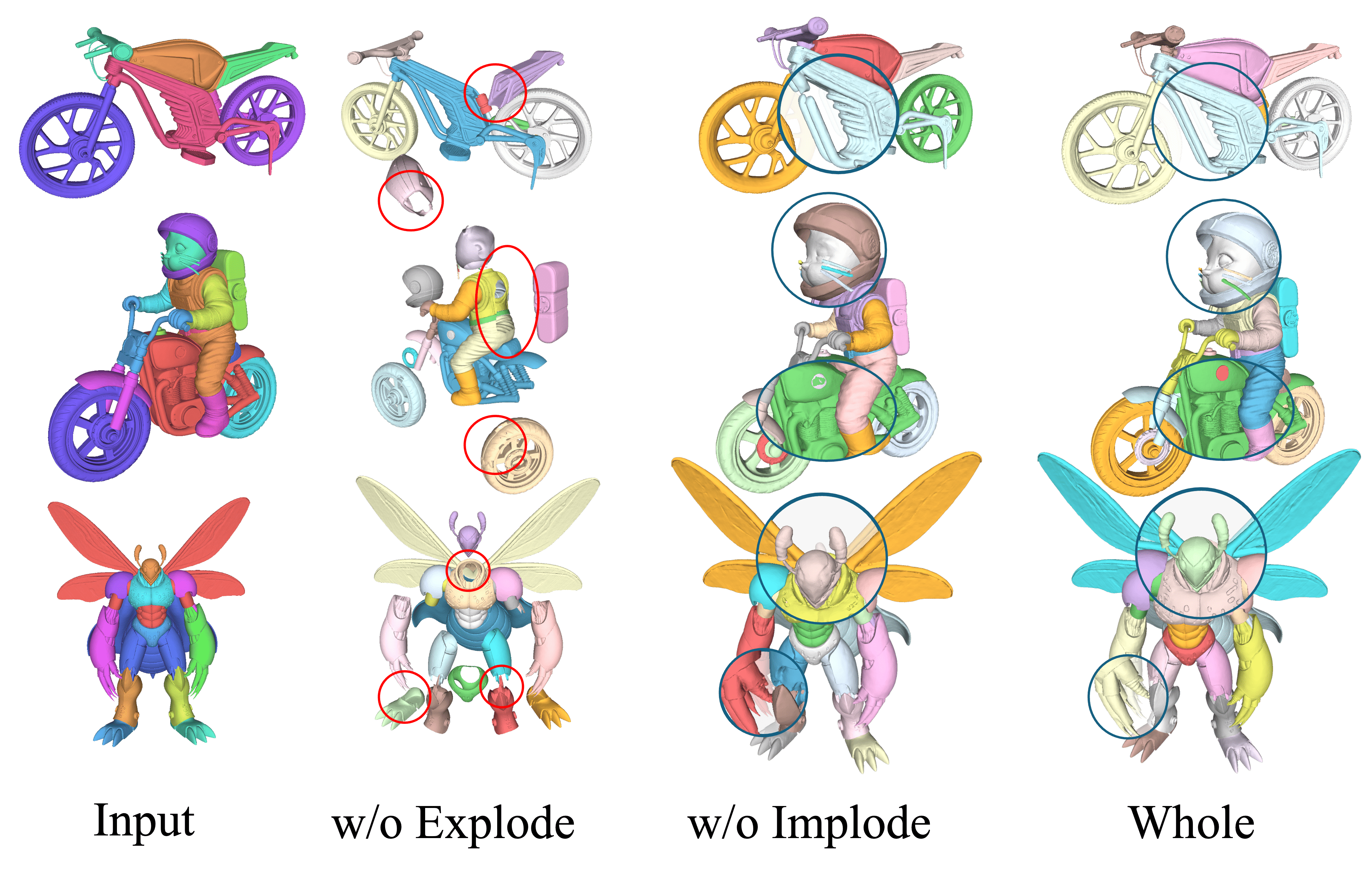}
  \caption{Qualitative comparison between our method and the alternative strategies. The input is the same segmentation.}
  \label{fig:abl_EI}
\end{figure}

%% file: figure/fig_ablation_PS.tex
\begin{figure}[htbp]
  \centering
  \includegraphics[width=0.91\linewidth]{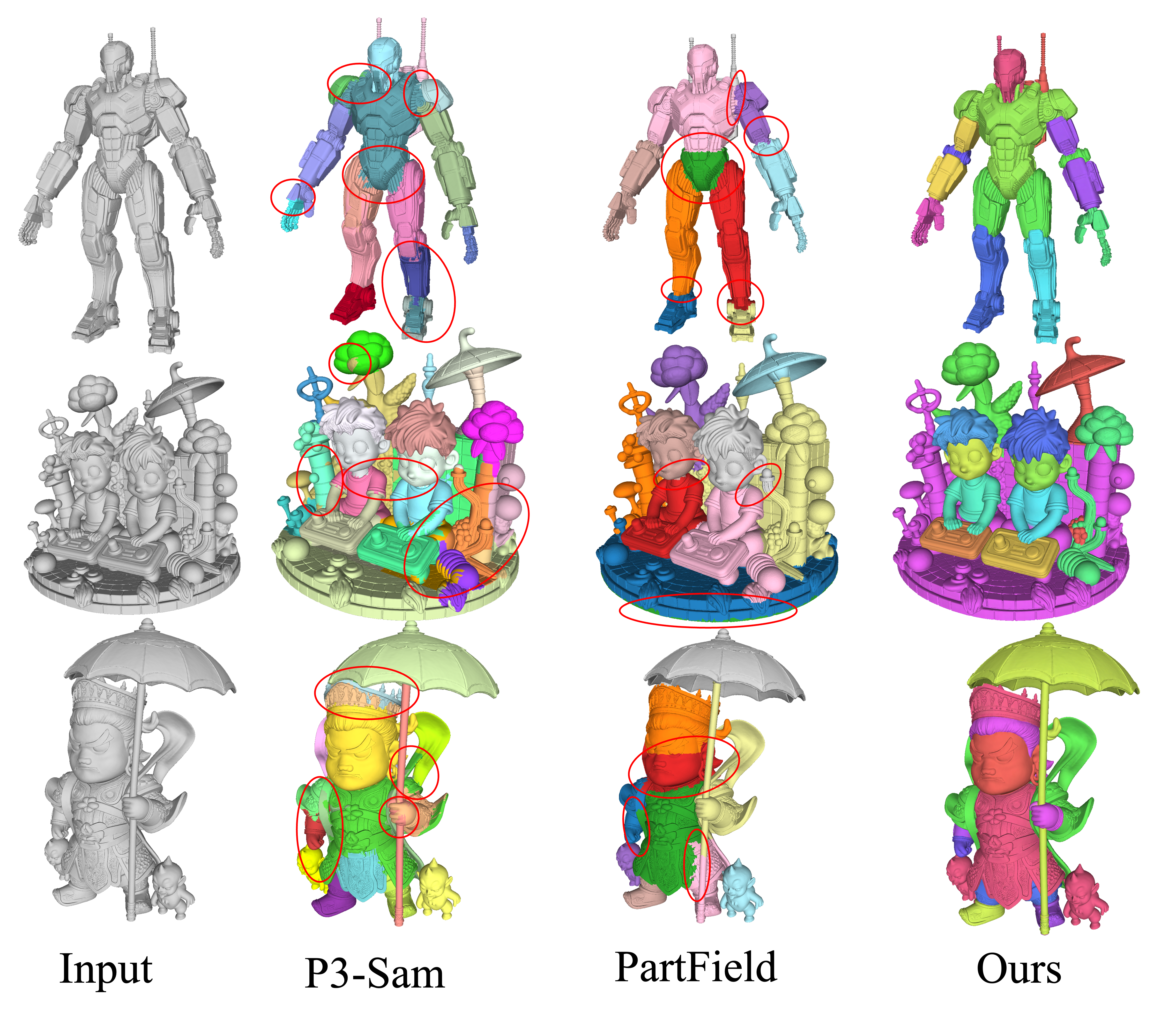}
  \caption{Qualitative comparison between our segmentation method  and the alternative methods.}
  \label{fig:abl_PS}
\end{figure}

%% file: section/05_conclusion.tex
\section{Conclusion}
\label{sec:conclusion}
Part-level generation remains a significant challenge in 3D content creation, as it requires balancing structural coherence and geometric plausibility while ensuring accuracy and computational efficiency. In this work, we present \methodName, a novel framework that addresses the critical challenges of 3D part generation through a novel Explode-Implode strategy, enabling the production of high-quality 3D shapes with well-constructed parts. The explosion phase allows for the comprehensive utilization of spatial capacity, facilitating meaningful part generation while mitigating issues related to overlapping voxels. Subsequently, the implosion phase maximizes spatial resolution, enabling the capture of fine-grained geometric details essential for realistic 3D assets. Our extensive experiments across various benchmarks demonstrate that \methodName outperforms existing state-of-the-art methods in part-level generation, yielding semantically meaningful and structurally coherent parts. The results underscore the effectiveness of our framework in producing high-quality 3D parts, thereby advancing the field of 3D content generation. Future work will explore further enhancements to our method by integrating physical principles into the generation process to better address complex scenarios that require high physical accuracy.

%% file: section/X_suppl.tex
\clearpage
\setcounter{page}{1}
\maketitlesupplementary
\input{figure/fig_compare_supp}
\input{table/table_ablation}

\section{Network Detatils}
\label{sec:network}
FluxDev-1~\cite{flux2024} is employed as the backbone of MVSegNet, while InSegNet adopts the large texturing architecture from UniTex~\cite{liang2025unitex}.

In the \textbf{Explode for Completion} stage, the DiT network utilizes the \textit{Sparse Latent Flow} architecture from Trellis~\cite{xiang2025structured}, which comprises 24 Transformer blocks. Since we adopt the Flux-Kontext~\cite{batifol2025flux} training strategy, a lightweight positional embedding is applied to distinguish noise latent tokens from conditional latent tokens. The resulting embeddings are added to the token features before being input into the DiT blocks.

For the \textbf{Implode for Refinement} stage, we utilize the Sparc3D~\cite{li2025sparc3d} VAE with a latent resolution of $64^3$. To avoid degradation in reconstruction quality, the downsampling operation in the original Stage-II Trellis~\cite{xiang2025structured} DiT network is removed, while the internal structure of each DiT block remains unchanged. Positional embeddings are consistently applied to both noise and conditional tokens to encode spatial positional information before being processed by the DiT blocks.

\section{Training Detatils}
\label{sec:training}
The training data pairs for the two-stage DiT are constructed as follows:

\begin{itemize}
\item \textbf{Incomplete part meshes}: These are obtained by first converting the complete mesh into a watertight representation, then discarding the internal structures while preserving the face correspondence information. The resulting meshes are used as the data for incomplete part meshes.
\item \textbf{Complete part meshes}: These are directly extracted from the original GLB files. Each individual part is processed to ensure watertightness and is then used as the ground-truth (GT) part mesh data.
\end{itemize}
In the first stage, the VAE is fine-tuned on the curated part dataset, and all mesh data are encoded into the latent space for DiT training. In the second stage, the original Sparc3D~\cite{li2025sparc3d} VAE is employed to encode meshes into latents, where the latents of incomplete sub-meshes serve as conditions, and the active voxel positions are guided by the GT part meshes.

\section{More Comparison}

\paragraph{Comparison with Baseline Methods.}
Fig.~\ref{fig:compare_supp} presents additional qualitative results compared to state-of-the-art (SOTA) methods. As illustrated, BANG~\cite{zhang2025bang} and PartPacker~\cite{tang2024partpacker} struggle to capture fine geometric details that correspond with the input model. This issue stems from their practice of embedding all components into a single or dual set of latent tokens, which limits the representational capacity of individual parts and ultimately leads to inaccuracies in geometric detail. HoloPart~\cite{yang2025holopart} neglects the interrelationships between parts, leading to unnatural completion outcomes that resemble merely filling a hole, thus compromising structural coherence. While PartCrafter~\cite{lin2025partcrafter} and X-Part~\cite{yan2025x} utilize a cross-attention mechanism to enhance structural coherence, their reliance on fixed-length part tokens proves inflexible for components of varying sizes. This limitation results in inefficient resource allocation, where larger components may be underrepresented while resources are wasted on smaller parts. OmniPart~\cite{yang2025omnipart} employs voxel-based representations to adaptively allocate spatial resolution to different parts; however, it supports completion only within active voxels, restricting the available spatial extents for completion. This constraint can lead to implausible part geometries and reduced geometric fidelity. In contrast, our method excels in generating high-quality 3D shapes with well-defined parts, demonstrating significant performance improvements in structural coherence, geometric plausibility, geometric fidelity, and efficiency.

\paragraph{Ablation Study on Explode and Implode.}
Tab.~\ref{tab:ablation} presents the quantitative comparisons between our method and alternative strategies. Our proposed method outperforms all the other options, achieving the highest scores.

%% file: figure/fig_compare_supp.tex
\begin{figure*}[htbp]
  \centering
  \includegraphics[width=1.\linewidth]{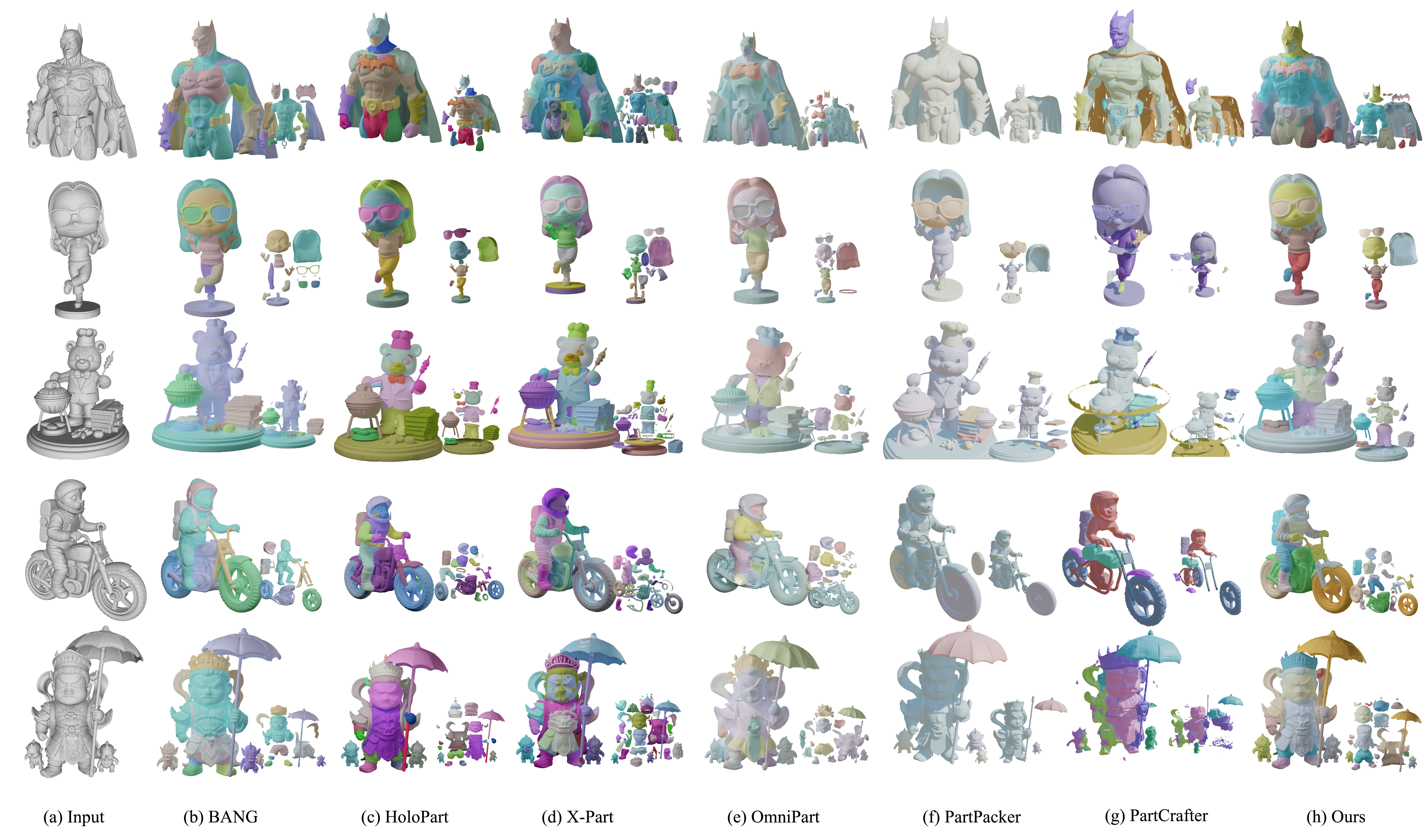}
  \caption{Qualitative comparison between ours and the state of the arts. For each row, the input model (a) is shown alongside the parts generated by (b) BANG~\cite{zhang2025bang}, (c) HoloPart~\cite{yang2025holopart}, (d) X-Part~\cite{yan2025x}, (e) OmniPart~\cite{yang2025omnipart}, (f) PartPacker~\cite{tang2024partpacker}, (g) PartCrafter~\cite{lin2025partcrafter}, and (h) our method. Our method demonstrates exceptional proficiency in generating high-quality 3D shapes with well-defined parts, achieving significant improvements in structural coherence, geometric plausibility, fidelity, and efficiency.}
  \label{fig:compare_supp}
\end{figure*}

%% file: table/table_ablation.tex
\begin{table*}[htbp]
    \centering
    \caption{Quantitative comparison across different settings at both the part and overall object levels. Since the part-level ground truth (GT) is derived from the complete results, the absence of the Explode operation results in lower part completion quality, underscoring its significance. Meanwhile, the explode–implode refinement slightly compromises overall mesh accuracy, revealing a trade-off between enhancing part-level details and maintaining global reconstruction precision.}
    \resizebox{1.\linewidth}{!}{
    \begin{tabular}{c|cccccc|cccccc}
        \multirow{2}{*}{Setting} & \multicolumn{6}{c|}{Part-level} & \multicolumn{6}{c}{Overall-object} \\
        \cline{2-13}
        & \textbf{Voxel IOU $\uparrow$} & \textbf{CD $\downarrow$} & \textbf{Voxel F-Score 0.01 $\uparrow$} & \textbf{F-Score 0.1 $\uparrow$} & \textbf{F-Score 0.05 $\uparrow$} & \textbf{F-Score 0.01 $\uparrow$} & \textbf{Voxel IOU $\uparrow$} & \textbf{CD $\downarrow$} & \textbf{Voxel F-Score 0.01 $\uparrow$} & \textbf{F-Score 0.1 $\uparrow$} & \textbf{F-Score 0.05 $\uparrow$} & \textbf{F-Score 0.01 $\uparrow$} \\
        \hline
        w/o Explode & 0.7647 & 0.0153 & 0.8883 & 0.9975 & 0.9866 & 0.7599 & \textbf{0.9310} & \textbf{0.0092} & \textbf{0.9642} & \textbf{1.0000} & \textbf{1.0000} & \textbf{0.9680} \\
        w/o Implode & 0.7992 & 0.0168 & 0.8664 & 0.9998 & 0.9937 & 0.8678 & 0.8036 & 0.0171 & 0.8896 & 0.9769 & 0.9615 & 0.8243 \\
        Whole       & -      & -      & -      & -      & -      & -      & 0.9005 & 0.0152 & 0.9091 & 0.9975 & 0.9869 & 0.8691 \\
    \end{tabular}
    }
    \label{tab:ablation}
\end{table*}

%% file: main.bib
@String(CVPR= {IEEE Conf. Comput. Vis. Pattern Recog.})

@String(ICCV= {Int. Conf. Comput. Vis.})

@String(ECCV= {Eur. Conf. Comput. Vis.})

@String(TOG= {ACM Trans. Graph.})

@String(CVPR  = {CVPR})

@String(ICCV  = {ICCV})

@String(ECCV  = {ECCV})

@String(TOG   = {ACM TOG})

@article{objaverse,
  title={Objaverse: A Universe of Annotated 3D Objects},
  author={Matt Deitke and Dustin Schwenk and Jordi Salvador and Luca Weihs and
          Oscar Michel and Eli VanderBilt and Ludwig Schmidt and
          Kiana Ehsani and Aniruddha Kembhavi and Ali Farhadi},
  journal={arXiv preprint arXiv:2212.08051},
  year={2022}
}

@article{objaverseXL,
  title={Objaverse-XL: A Universe of 10M+ 3D Objects},
  author={Matt Deitke and Ruoshi Liu and Matthew Wallingford and Huong Ngo and
          Oscar Michel and Aditya Kusupati and Alan Fan and Christian Laforte and
          Vikram Voleti and Samir Yitzhak Gadre and Eli VanderBilt and
          Aniruddha Kembhavi and Carl Vondrick and Georgia Gkioxari and
          Kiana Ehsani and Ludwig Schmidt and Ali Farhadi},
  journal={arXiv preprint arXiv:2307.05663},
  year={2023}
}

@article{zhang2024clay,
  title={Clay: A controllable large-scale generative model for creating high-quality 3d assets},
  author={Zhang, Longwen and Wang, Ziyu and Zhang, Qixuan and Qiu, Qiwei and Pang, Anqi and Jiang, Haoran and Yang, Wei and Xu, Lan and Yu, Jingyi},
  journal={ACM Transactions on Graphics (TOG)},
  volume={43},
  number={4},
  pages={1--20},
  year={2024},
  publisher={ACM New York, NY, USA}
}

@article{li2024craftsman3d,
  title={Craftsman3d: High-fidelity mesh generation with 3d native generation and interactive geometry refiner},
  author={Li, Weiyu and Liu, Jiarui and Yan, Hongyu and Chen, Rui and Liang, Yixun and Chen, Xuelin and Tan, Ping and Long, Xiaoxiao},
  journal={arXiv preprint arXiv:2405.14979},
  year={2024}
}

@inproceedings{rombach2022high,
  title={High-resolution image synthesis with latent diffusion models},
  author={Rombach, Robin and Blattmann, Andreas and Lorenz, Dominik and Esser, Patrick and Ommer, Bj{\"o}rn},
  booktitle={Proceedings of the IEEE/CVF conference on computer vision and pattern recognition},
  pages={10684--10695},
  year={2022}
}

@inproceedings{liu2023zero,
  title={Zero-1-to-3: Zero-shot one image to 3d object},
  author={Liu, Ruoshi and Wu, Rundi and Van Hoorick, Basile and Tokmakov, Pavel and Zakharov, Sergey and Vondrick, Carl},
  booktitle={Proceedings of the IEEE/CVF international conference on computer vision},
  pages={9298--9309},
  year={2023}
}

@inproceedings{xiang2025structured,
  title={Structured 3d latents for scalable and versatile 3d generation},
  author={Xiang, Jianfeng and Lv, Zelong and Xu, Sicheng and Deng, Yu and Wang, Ruicheng and Zhang, Bowen and Chen, Dong and Tong, Xin and Yang, Jiaolong},
  booktitle={Proceedings of the Computer Vision and Pattern Recognition Conference},
  pages={21469--21480},
  year={2025}
}

@article{li2025triposg,
  title={Triposg: High-fidelity 3d shape synthesis using large-scale rectified flow models},
  author={Li, Yangguang and Zou, Zi-Xin and Liu, Zexiang and Wang, Dehu and Liang, Yuan and Yu, Zhipeng and Liu, Xingchao and Guo, Yuan-Chen and Liang, Ding and Ouyang, Wanli and others},
  journal={arXiv preprint arXiv:2502.06608},
  year={2025}
}

@article{zhao2025hunyuan3d,
  title={Hunyuan3d 2.0: Scaling diffusion models for high resolution textured 3d assets generation},
  author={Zhao, Zibo and Lai, Zeqiang and Lin, Qingxiang and Zhao, Yunfei and Liu, Haolin and Yang, Shuhui and Feng, Yifei and Yang, Mingxin and Zhang, Sheng and Yang, Xianghui and others},
  journal={arXiv preprint arXiv:2501.12202},
  year={2025}
}

@article{chen2025ultra3d,
  title={Ultra3d: Efficient and high-fidelity 3d generation with part attention},
  author={Chen, Yiwen and Li, Zhihao and Wang, Yikai and Zhang, Hu and Li, Qin and Zhang, Chi and Lin, Guosheng},
  journal={arXiv preprint arXiv:2507.17745},
  year={2025}
}

@article{ye2025hi3dgen,
  title={Hi3dgen: High-fidelity 3d geometry generation from images via normal bridging},
  author={Ye, Chongjie and Wu, Yushuang and Lu, Ziteng and Chang, Jiahao and Guo, Xiaoyang and Zhou, Jiaqing and Zhao, Hao and Han, Xiaoguang},
  journal={arXiv preprint arXiv:2503.22236},
  volume={3},
  pages={2},
  year={2025}
}

@article{li2025sparc3d,
  title={Sparc3D: Sparse Representation and Construction for High-Resolution 3D Shapes Modeling},
  author={Li, Zhihao and Wang, Yufei and Zheng, Heliang and Luo, Yihao and Wen, Bihan},
  journal={arXiv preprint arXiv:2505.14521},
  year={2025}
}

@article{yang2025holopart,
  title={Holopart: Generative 3d part amodal segmentation},
  author={Yang, Yunhan and Guo, Yuan-Chen and Huang, Yukun and Zou, Zi-Xin and Yu, Zhipeng and Li, Yangguang and Cao, Yan-Pei and Liu, Xihui},
  journal={arXiv preprint arXiv:2504.07943},
  year={2025}
}

@article{zhang2025bang,
  title={BANG: Dividing 3D Assets via Generative Exploded Dynamics},
  author={Zhang, Longwen and Zhang, Qixuan and Jiang, Haoran and Bai, Yinuo and Yang, Wei and Xu, Lan and Yu, Jingyi},
  journal={ACM Transactions on Graphics (TOG)},
  volume={44},
  number={4},
  pages={1--21},
  year={2025},
  publisher={ACM New York, NY, USA}
}

@article{lin2025partcrafter,
  title={PartCrafter: Structured 3D Mesh Generation via Compositional Latent Diffusion Transformers},
  author={Lin, Yuchen and Lin, Chenguo and Pan, Panwang and Yan, Honglei and Feng, Yiqiang and Mu, Yadong and Fragkiadaki, Katerina},
  journal={arXiv preprint arXiv:2506.05573},
  year={2025}
}

@article{tang2024partpacker,
  title={Efficient Part-level 3D Object Generation via Dual Volume Packing},
  author={Tang, Jiaxiang and Lu, Ruijie and Li, Zhaoshuo and Hao, Zekun and Li, Xuan and Wei, Fangyin and Song, Shuran and Zeng, Gang and Liu, Ming-Yu and Lin, Tsung-Yi},
  journal={arXiv preprint arXiv:2506.09980},
  year={2025}
}

@article{yan2025x,
  title={X-Part: high fidelity and structure coherent shape decomposition},
  author={Yan, Xinhao and Xu, Jiachen and Li, Yang and Ma, Changfeng and Yang, Yunhan and Wang, Chunshi and Zhao, Zibo and Lai, Zeqiang and Zhao, Yunfei and Chen, Zhuo and others},
  journal={arXiv preprint arXiv:2509.08643},
  year={2025}
}

@article{yang2025omnipart,
  title={Omnipart: Part-aware 3d generation with semantic decoupling and structural cohesion},
  author={Yang, Yunhan and Zhou, Yufan and Guo, Yuan-Chen and Zou, Zi-Xin and Huang, Yukun and Liu, Ying-Tian and Xu, Hao and Liang, Ding and Cao, Yan-Pei and Liu, Xihui},
  journal={arXiv preprint arXiv:2507.06165},
  year={2025}
}

@article{chang2015shapenet,
  title={Shapenet: An information-rich 3d model repository},
  author={Chang, Angel X and Funkhouser, Thomas and Guibas, Leonidas and Hanrahan, Pat and Huang, Qixing and Li, Zimo and Savarese, Silvio and Savva, Manolis and Song, Shuran and Su, Hao and others},
  journal={arXiv preprint arXiv:1512.03012},
  year={2015}
}

@inproceedings{wang2018pixel2mesh,
  title={Pixel2mesh: Generating 3d mesh models from single rgb images},
  author={Wang, Nanyang and Zhang, Yinda and Li, Zhuwen and Fu, Yanwei and Liu, Wei and Jiang, Yu-Gang},
  booktitle={Proceedings of the European conference on computer vision (ECCV)},
  pages={52--67},
  year={2018}
}

@inproceedings{park2019deepsdf,
  title={Deepsdf: Learning continuous signed distance functions for shape representation},
  author={Park, Jeong Joon and Florence, Peter and Straub, Julian and Newcombe, Richard and Lovegrove, Steven},
  booktitle={Proceedings of the IEEE/CVF conference on computer vision and pattern recognition},
  pages={165--174},
  year={2019}
}

@inproceedings{mescheder2019occupancy,
  title={Occupancy networks: Learning 3d reconstruction in function space},
  author={Mescheder, Lars and Oechsle, Michael and Niemeyer, Michael and Nowozin, Sebastian and Geiger, Andreas},
  booktitle={Proceedings of the IEEE/CVF conference on computer vision and pattern recognition},
  pages={4460--4470},
  year={2019}
}

@inproceedings{saito2019pifu,
  title={Pifu: Pixel-aligned implicit function for high-resolution clothed human digitization},
  author={Saito, Shunsuke and Huang, Zeng and Natsume, Ryota and Morishima, Shigeo and Kanazawa, Angjoo and Li, Hao},
  booktitle={Proceedings of the IEEE/CVF international conference on computer vision},
  pages={2304--2314},
  year={2019}
}

@article{luo2023sketchmetaface,
  title={SketchMetaFace: A learning-based sketching interface for high-fidelity 3D character face modeling},
  author={Luo, Zhongjin and Du, Dong and Zhu, Heming and Yu, Yizhou and Fu, Hongbo and Han, Xiaoguang},
  journal={IEEE Transactions on Visualization and Computer Graphics},
  volume={30},
  number={8},
  pages={5260--5275},
  year={2023},
  publisher={IEEE}
}

@article{zhang20233dshape2vecset,
  title={3dshape2vecset: A 3d shape representation for neural fields and generative diffusion models},
  author={Zhang, Biao and Tang, Jiapeng and Niessner, Matthias and Wonka, Peter},
  journal={ACM Transactions On Graphics (TOG)},
  volume={42},
  number={4},
  pages={1--16},
  year={2023},
  publisher={ACM New York, NY, USA}
}

@inproceedings{shue20233d,
  title={3d neural field generation using triplane diffusion},
  author={Shue, J Ryan and Chan, Eric Ryan and Po, Ryan and Ankner, Zachary and Wu, Jiajun and Wetzstein, Gordon},
  booktitle={Proceedings of the IEEE/CVF Conference on Computer Vision and Pattern Recognition},
  pages={20875--20886},
  year={2023}
}

@inproceedings{ren2024xcube,
  title={Xcube: Large-scale 3d generative modeling using sparse voxel hierarchies},
  author={Ren, Xuanchi and Huang, Jiahui and Zeng, Xiaohui and Museth, Ken and Fidler, Sanja and Williams, Francis},
  booktitle={Proceedings of the IEEE/CVF conference on computer vision and pattern recognition},
  pages={4209--4219},
  year={2024}
}

@inproceedings{li2023diffusion,
  title={Diffusion-sdf: Text-to-shape via voxelized diffusion},
  author={Li, Muheng and Duan, Yueqi and Zhou, Jie and Lu, Jiwen},
  booktitle={Proceedings of the IEEE/CVF conference on computer vision and pattern recognition},
  pages={12642--12651},
  year={2023}
}

@article{kerbl20233d,
  title={3D Gaussian splatting for real-time radiance field rendering.},
  author={Kerbl, Bernhard and Kopanas, Georgios and Leimk{\"u}hler, Thomas and Drettakis, George},
  journal={ACM Trans. Graph.},
  volume={42},
  number={4},
  pages={139--1},
  year={2023}
}

@inproceedings{siddiqui2024meshgpt,
  title={Meshgpt: Generating triangle meshes with decoder-only transformers},
  author={Siddiqui, Yawar and Alliegro, Antonio and Artemov, Alexey and Tommasi, Tatiana and Sirigatti, Daniele and Rosov, Vladislav and Dai, Angela and Nie{\ss}ner, Matthias},
  booktitle={Proceedings of the IEEE/CVF conference on computer vision and pattern recognition},
  pages={19615--19625},
  year={2024}
}

@article{chen2024meshanything,
  title={Meshanything: Artist-created mesh generation with autoregressive transformers},
  author={Chen, Yiwen and He, Tong and Huang, Di and Ye, Weicai and Chen, Sijin and Tang, Jiaxiang and Chen, Xin and Cai, Zhongang and Yang, Lei and Yu, Gang and others},
  journal={arXiv preprint arXiv:2406.10163},
  year={2024}
}

@article{hao2024meshtron,
  title={Meshtron: High-fidelity, artist-like 3d mesh generation at scale},
  author={Hao, Zekun and Romero, David W and Lin, Tsung-Yi and Liu, Ming-Yu},
  journal={arXiv preprint arXiv:2412.09548},
  year={2024}
}

@article{wang2025nautilus,
  title={Nautilus: Locality-aware Autoencoder for Scalable Mesh Generation},
  author={Wang, Yuxuan and Yi, Xuanyu and Weng, Haohan and Xu, Qingshan and Wei, Xiaokang and Yang, Xianghui and Guo, Chunchao and Chen, Long and Zhang, Hanwang},
  journal={arXiv preprint arXiv:2501.14317},
  year={2025}
}

@article{chen2024partgen,
  title={PartGen: Part-level 3D Generation and Reconstruction with Multi-View Diffusion Models},
  author={Minghao Chen and Roman Shapovalov and Iro Laina and Tom Monnier and Jianyuan Wang and David Novotny and Andrea Vedaldi},
  journal={arXiv preprint arXiv:2412.18608},
  year={2024}
}

@inproceedings{qi2017pointnet,
  title={Pointnet: Deep learning on point sets for 3d classification and segmentation},
  author={Qi, Charles R and Su, Hao and Mo, Kaichun and Guibas, Leonidas J},
  booktitle={Proceedings of the IEEE conference on computer vision and pattern recognition},
  pages={652--660},
  year={2017}
}

@article{qi2017pointnet++,
  title={Pointnet++: Deep hierarchical feature learning on point sets in a metric space},
  author={Qi, Charles Ruizhongtai and Yi, Li and Su, Hao and Guibas, Leonidas J},
  journal={Advances in neural information processing systems},
  volume={30},
  year={2017}
}

@inproceedings{zhao2021point,
  title={Point transformer},
  author={Zhao, Hengshuang and Jiang, Li and Jia, Jiaya and Torr, Philip HS and Koltun, Vladlen},
  booktitle={Proceedings of the IEEE/CVF international conference on computer vision},
  pages={16259--16268},
  year={2021}
}

@article{kirillov2023segany,
  title={Segment Anything},
  author={Kirillov, Alexander and Mintun, Eric and Ravi, Nikhila and Mao, Hanzi and Rolland, Chloe and Gustafson, Laura and Xiao, Tete and Whitehead, Spencer and Berg, Alexander C. and Lo, Wan-Yen and Doll{\'a}r, Piotr and Girshick, Ross},
  journal={arXiv:2304.02643},
  year={2023}
}

@article{hafner2021clip,
  title={CLIP and complementary methods},
  author={Hafner, Markus and Katsantoni, Maria and K{\"o}ster, Tino and Marks, James and Mukherjee, Joyita and Staiger, Dorothee and Ule, Jernej and Zavolan, Mihaela},
  journal={Nature Reviews Methods Primers},
  volume={1},
  number={1},
  pages={20},
  year={2021},
  publisher={Nature Publishing Group UK London}
}

@inproceedings{caron2021emerging,
  title={Emerging Properties in Self-Supervised Vision Transformers},
  author={Caron, Mathilde and Touvron, Hugo and Misra, Ishan and J\'egou, Herv\'e  and Mairal, Julien and Bojanowski, Piotr and Joulin, Armand},
  booktitle={Proceedings of the International Conference on Computer Vision (ICCV)},
  year={2021}
}

@inproceedings{zhong2024meshsegmenter,
  title={Meshsegmenter: Zero-shot mesh semantic segmentation via texture synthesis},
  author={Zhong, Ziming and Xu, Yanyu and Li, Jing and Xu, Jiale and Li, Zhengxin and Yu, Chaohui and Gao, Shenghua},
  booktitle={European Conference on Computer Vision},
  pages={182--199},
  year={2024},
  organization={Springer}
}

@inproceedings{umam2023partdistill,
    title = {PartDistill: 3D Shape Part Segmentation by Vision-Language Model Distillation},
    author = {Umam, Ardian and Yang, Cheng-Kun and Chen, Min-Hung and Chuang, Jen-Hui and Lin, Yen-Yu},
    booktitle = {IEEE/CVF International Conference on Computer Vision (CVPR)},
    year = {2024},
}

@inproceedings{xu2025sampro3d,
    title={SAMPro3D: Locating SAM Prompts in 3D for Zero-Shot Instance Segmentation}, 
    author={Mutian Xu and Xingyilang Yin and Lingteng Qiu and Yang Liu and Xin Tong and Xiaoguang Han},
    year={2025},
    booktitle = {International Conference on 3D Vision (3DV)}
  }

@article{tang2024segment,
  title={Segment any mesh: Zero-shot mesh part segmentation via lifting segment anything 2 to 3d},
  author={Tang, George and Zhao, William and Ford, Logan and Benhaim, David and Zhang, Paul},
  journal={arXiv e-prints},
  pages={arXiv--2408},
  year={2024}
}

@article{yang2024sampart3d,
  title={Sampart3d: Segment any part in 3d objects},
  author={Yang, Yunhan and Huang, Yukun and Guo, Yuan-Chen and Lu, Liangjun and Wu, Xiaoyang and Lam, Edmund Y and Cao, Yan-Pei and Liu, Xihui},
  journal={arXiv preprint arXiv:2411.07184},
  year={2024}
}

@inproceedings{liu2025partfield,
  title={Partfield: Learning 3d feature fields for part segmentation and beyond},
  author={Liu, Minghua and Uy, Mikaela Angelina and Xiang, Donglai and Su, Hao and Fidler, Sanja and Sharp, Nicholas and Gao, Jun},
  booktitle={Proceedings of the IEEE/CVF International Conference on Computer Vision},
  pages={9704--9715},
  year={2025}
}

@article{ma2025p3,
  title={P3-sam: Native 3d part segmentation},
  author={Ma, Changfeng and Li, Yang and Yan, Xinhao and Xu, Jiachen and Yang, Yunhan and Wang, Chunshi and Zhao, Zibo and Guo, Yanwen and Chen, Zhuo and Guo, Chunchao},
  journal={arXiv preprint arXiv:2509.06784},
  year={2025}
}

@article{bensadoun2024meta,
  title={Meta 3d texturegen: Fast and consistent texture generation for 3d objects},
  author={Bensadoun, Raphael and Kleiman, Yanir and Azuri, Idan and Harosh, Omri and Vedaldi, Andrea and Neverova, Natalia and Gafni, Oran},
  journal={arXiv preprint arXiv:2407.02430},
  year={2024}
}

@article{yang2025pandora3d,
  title={Pandora3d: A comprehensive framework for high-quality 3d shape and texture generation},
  author={Yang, Jiayu and Shang, Taizhang and Sun, Weixuan and Song, Xibin and Cheng, Ziang and Wang, Senbo and Chen, Shenzhou and Liu, Weizhe and Li, Hongdong and Ji, Pan},
  journal={arXiv preprint arXiv:2502.14247},
  year={2025}
}

@article{liang2025unitex,
  title={UniTEX: Universal High Fidelity Generative Texturing for 3D Shapes},
  author={Liang, Yixun and Luo, Kunming and Chen, Xiao and Chen, Rui and Yan, Hongyu and Li, Weiyu and Liu, Jiarui and Tan, Ping},
  journal={arXiv preprint arXiv:2505.23253},
  year={2025}
}

@article{batifol2025flux,
  title={FLUX. 1 Kontext: Flow Matching for In-Context Image Generation and Editing in Latent Space},
  author={Batifol, Stephen and Blattmann, Andreas and Boesel, Frederic and Consul, Saksham and Diagne, Cyril and Dockhorn, Tim and English, Jack and English, Zion and Esser, Patrick and Kulal, Sumith and others},
  journal={arXiv e-prints},
  pages={arXiv--2506},
  year={2025}
}

@article{lipman2022flow,
  title={Flow matching for generative modeling},
  author={Lipman, Yaron and Chen, Ricky TQ and Ben-Hamu, Heli and Nickel, Maximilian and Le, Matt},
  journal={arXiv preprint arXiv:2210.02747},
  year={2022}
}

@inproceedings{peebles2023scalable,
  title={Scalable diffusion models with transformers},
  author={Peebles, William and Xie, Saining},
  booktitle={Proceedings of the IEEE/CVF international conference on computer vision},
  pages={4195--4205},
  year={2023}
}

@article{oquab2023dinov2,
  title={Dinov2: Learning robust visual features without supervision},
  author={Oquab, Maxime and Darcet, Timoth{\'e}e and Moutakanni, Th{\'e}o and Vo, Huy and Szafraniec, Marc and Khalidov, Vasil and Fernandez, Pierre and Haziza, Daniel and Massa, Francisco and El-Nouby, Alaaeldin and others},
  journal={arXiv preprint arXiv:2304.07193},
  year={2023}
}

@inproceedings{collins2022abo,
  title={Abo: Dataset and benchmarks for real-world 3d object understanding},
  author={Collins, Jasmine and Goel, Shubham and Deng, Kenan and Luthra, Achleshwar and Xu, Leon and Gundogdu, Erhan and Zhang, Xi and Vicente, Tomas F Yago and Dideriksen, Thomas and Arora, Himanshu and others},
  booktitle={Proceedings of the IEEE/CVF conference on computer vision and pattern recognition},
  pages={21126--21136},
  year={2022}
}

@article{fu20213d,
  title={3d-future: 3d furniture shape with texture},
  author={Fu, Huan and Jia, Rongfei and Gao, Lin and Gong, Mingming and Zhao, Binqiang and Maybank, Steve and Tao, Dacheng},
  journal={International Journal of Computer Vision},
  volume={129},
  number={12},
  pages={3313--3337},
  year={2021},
  publisher={Springer}
}

@inproceedings{khanna2024habitat,
  title={Habitat synthetic scenes dataset (hssd-200): An analysis of 3d scene scale and realism tradeoffs for objectgoal navigation},
  author={Khanna, Mukul and Mao, Yongsen and Jiang, Hanxiao and Haresh, Sanjay and Shacklett, Brennan and Batra, Dhruv and Clegg, Alexander and Undersander, Eric and Chang, Angel X and Savva, Manolis},
  booktitle={Proceedings of the IEEE/CVF Conference on Computer Vision and Pattern Recognition},
  pages={16384--16393},
  year={2024}
}

@article{dong2025copart,
  title={From One to More: Contextual Part Latents for 3D Generation},
author={Shaocong Dong and Lihe Ding and Xiao Chen and Yaokun Li and Yuxin WANG and Yucheng Wang and Qi WANG and Jaehyeok Kim and Chenjian Gao and Zhanpeng Huang and Zibin Wang and Tianfan Xue and Dan Xu},
  booktitle={ICCV},
  year={2025}
}

@misc{flux2024,
    author={Black Forest Labs},
    title={FLUX},
    year={2024},
    howpublished={\url{https://github.com/black-forest-labs/flux}},
}
